\documentclass[11pt]{article}
\usepackage[dvips]{graphicx}
\usepackage{amssymb}
\usepackage{natbib}
\usepackage[headings]{fullpage}
\def\etal{\emph{et al.}}

\begin{document}
\title{Bayesian History Reconstruction of 
       Complex Human Gene Clusters
       on a Phylogeny}
\author{Tom\'a\v{s} Vina\v{r}$^1$,
        Bro\v{n}a Brejov\'a$^1$, 
        Giltae Song$^2$, and
        Adam Siepel$^3$}

\date{$^1$ Faculty of Mathematics, Physics and Informatics, Comenius University,
Mlynsk\'a Dolina, 842~48 Bratislava, Slovakia\\
$^2$ Center for Comparative Genomics and Bioinformatics, 506B Wartik Lab, Penn State University, University Park, PA 16802, USA\\
$^3$ Dept. of Biological Statistics and Comp. Biology,
Cornell University, Ithaca, NY 14853, USA\\}

\maketitle

\begin{abstract}
Clusters of genes that have evolved by repeated segmental duplication
present difficult challenges throughout genomic
analysis, from sequence assembly to functional analysis.  Improved understanding
of these clusters is of utmost importance, since they have been shown
to be the source of evolutionary innovation, and have been linked to
multiple diseases, including HIV and a variety of cancers. Previously,
Zhang \etal~(2008)  developed an algorithm for reconstructing
parsimonious evolutionary histories of such gene clusters, using only
human genomic sequence data. In this paper, we propose a
probabilistic model for the evolution of gene clusters on a phylogeny,
and an MCMC algorithm for reconstruction of duplication histories
from genomic sequences in multiple species. Several projects are underway
to obtain high quality BAC-based assemblies of 
duplicated clusters in multiple species, and we anticipate that our method 
will be useful in analyzing these valuable new data sets.
\end{abstract}

\section{Introduction}

Segmental duplications cover about 5\% of the human genome
\citep{Lander2001}.  When multiple segmental duplications occur
at a particular genomic locus they give rise to complex gene
clusters. Many important gene families 
linked to various diseases, including cancers, Alzheimer's disease, and
HIV, reside 
in such clusters. Gene
duplication is often followed by functional diversification
\citep{Ohno1970}, and, indeed, genes overlapping segmental duplications
have been shown to be enriched for positive selection
\citep{Gibbs2007}. 

In this paper, we describe a probabilistic model of evolution of such
gene clusters on a phylogeny, and devise a Markov-chain Monte Carlo
sampling algorithm for inference of highly probable duplication
histories and ancestral sequences. To demonstrate
the usefulness of our approach, we apply our algorithm to simulated
sequences on human-chimp-macaque phylogeny, as well as to real clusters
assembled from available BAC sequencing data.

Previously, 
\citet{Elemento2002,Lajoie2007} studied the reconstruction of gene
family histories by considering tandem duplications and inversions
as the only possible events. They also assume that 
genes are always copied as a whole unit. \citet{Zhang2008}
demonstrated that 
more complex
models are needed to address evolution of gene clusters in the human
genome.

\begin{figure}
\centerline{\includegraphics[width=0.4\columnwidth]{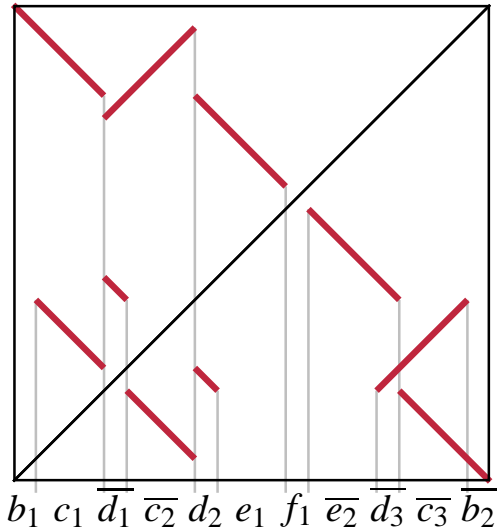}}%
\caption{{\bf Sequence atomization.}
Simulated self-alignment of a result of three duplication
events. Lines represent local sequence alignments. There are five
types of atomic segments ($b,c,d,e,f$). For example, type $d$ has
three copies: one on the forwards strand ($d_2$) and two on the
reverse strand ($\overline{d_1},\overline{d_3}$).
\label{fig:atomization}}
\end{figure}

In more recent work, genes have
been replaced by generic \emph{atomic segments} \citep{Zhang2008,Ma2008} as
the substrates of reconstruction algorithms.
Briefly, a self-alignment is constructed by a local alignment program
(e.g., blastz \citep{Schwartz2003}), and only alignments above
certain threshold (e.g., 93\% for human-macaque split) are kept.  The
boundaries of alignments mark \emph{breakpoints}, and
the sequences between neighboring breakpoints 
are considered atomic segments
(Fig.\ref{fig:atomization}).
Due to the \emph{transitivity} of sequence similarity between atomic
segments, the set of atomic segments can be decomposed
into equivalence classes, or \emph{atom types}. Thus, the
nucleotide sequence is transformed into a simpler sequence of
atoms.

The task of \emph{duplication history reconstruction} is to find
a sequence of evolutionary events (e.g., duplications, deletions, 
and speciations) that starts with an ancestral sequence of atoms, 
in which no atom type occurs twice, and ends
with atomic sequences of extant species. Such a history also
directly implies ``gene trees'' of individual atomic types,
which we call \emph{segment trees}. These trees
are implicitly rooted and reconciled with the species tree, and this
information can be easily used to reconstruct ancestral sequences at
speciation points segment by segment (see e.g.
\citep{Blanchette2004r}). A common way of looking at these histories
is from the most recent events back in time. In this context,
we can start from extant sequences, and \emph{unwind} events one-by-one,
until the ancestral sequence is reached.

\citet{Zhang2008} sought solutions of this problems 
with small number of events, 
given the sequence from a single species. In particular, they
proved a necessary condition to identify candidates
for the latest duplication operation, assuming no reuse of breakpoints. 
After unwinding the
latest duplication, the same step is
repeated to identify the second latest duplication, etc. 
Zhang \etal{} showed that following any sequence of 
candidate duplications leads to a history with the same number of
duplication events under no-breakpoint-reuse assumption. 
As a result, there may be an exponential number of
most parsimonious solutions to the problem, and it may be
impossible to reconstruct a unique history. 

A similar parsimony problem has also been recently explored by 
\citet{Ma2008} in the context of much larger sequences (whole
genomes) and a broader set of operations (including inversions,
translocations, etc.). In their algorithm, Ma \etal{} reconstruct
phylogenetic trees for every atomic segment, and reconcile these
segment trees with the species tree to infer deletions and
rooting. The authors give a polynomial-time algorithm for the history
reconstruction, assuming no-breakpoint-reuse and correct atomic
segment trees. Both methods make use of fairly extensive heuristics to
overcome violations of their assumptions and 
allow their algorithms to be applied to real data.

\begin{figure}
\centerline{
\includegraphics[width=0.6\columnwidth]{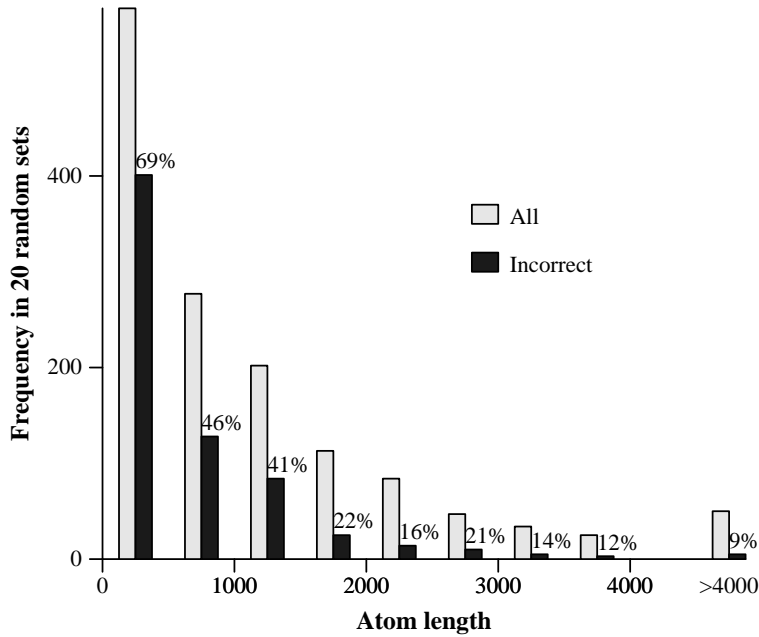}}
\caption{{\bf Distribution of atomic segment lengths and accuracy 
of segment tree inference}
 in 20 simulated fast-evolving clusters (see Section
\ref{sec:simulations}). The gray bars show the numbers of segment types.
The black bars show the percentages of segment types for which the
highest posterior probability unrooted segment tree inferred by MrBayes
\citep{Ronquist2003} does not match the correct segment tree.}
\label{fig:treeaccuracy}
\end{figure}

The no-breakpoint-reuse assumption is often justified by the
argument that in long sequences, it is unlikely that the same
breakpoint is used twice \citep{Nadeau1984}. 
However, there is evidence that breakpoints do not occur uniformly
throughout the sequence, and that breakpoint reuse is frequent
\citep{Peng2006,Becker2007}. Moreover, breakpoints located close to
each other may lead to short atoms that 
can't be reliably identified by sequence similarity
algorithms and categorized into atom types. For example,
in our simulated data (Section \ref{sec:simulations}), 
approximately 2\% of atoms are shorter than 20bp and may appear 
as additional breakpoint reuses instead. Thus,
no-breakpoint-reuse can be a useful guide, but cannot be
entirely relied on in application to real data sets. 
We have also examined the assumption of correctness of segment trees 
inferred from sequences of individual segments 
(Fig.\ref{fig:treeaccuracy}). For segments
shorter than 500bp (39\% of all segments in our simulations) 
69\% of the trees were incorrectly reconstructed,
and even for segments 500-1,000bp long, a substantial fraction is incorrect
(46\%).

In this paper, we present a simple probabilistic model for sequence
evolution by duplication, and we design a sampling algorithm that
explicitly accounts for uncertainty in the estimation of segment trees
and allows for breakpoint reuse. The results of
\citet{Zhang2008} suggest that, in spite of an improved
model, there may still be many solutions of similar
likelihood. The stochastic sampling approach allows us to
examine such multiple solutions in the same framework and extract
expectations for quantities of particular interest (e.g., the expected
number of events on individual branches of the phylogeny, or local
properties of the ancestral sequences). 
In addition, by using data from multiple species, our approach obtains
additional information about ancestral configurations.

Our problem is closely related to the problem of reconstruction of
gene trees and their reconciliation with species trees. Recent
algorithms for gene tree reconstruction (e.g., \citet{Wapinski2007})
also consider genomic context of individual genes. However, our
algorithms for reconstruction of duplication histories not only use
such context as an additional piece of information, but the derived
evolutionary histories also explain how similarities in the
genomic context of individual genes evolved.

Our current approach uses a simple
HKY nucleotide substitution model \citep{Hasegawa1985}, 
with variance in rates allowed
between individual atomic segments. However, in future work it will be
possible to employ more complex models of sequence evolution, such as
variable rate site models and models of codon evolution, within the
same framework. Such extensions will allow us to 
identify sites and branches under selection in
gene clusters in a principled way, and contribute towards better functional
characterization of these important genomic regions.

\section{Probabilistic model of evolution with segmental duplication}

In this section, we give a probabilistic model 
through segmental duplication on a given species tree $T$.
Such a model can be used to generate simulated data, as well as for
inference.

We start with an ancestral sequence of length $N$. In our model, this
sequence evolves by duplications, deletions and substitutions. 
A \emph{duplication}
copies a source region and inserts the new copy at a target position in
the sequence, either on the forward strand (with
probability $1-P_i$) or on the reverse strand (with probability $P_i$).  
A duplication can be characterized by
four coordinates: a \emph{centroid} (the midpoint of the region between the
leftmost and rightmost end of the duplication), the \emph{length} of the
source region, the \emph{distance} between the source and the target, and
a \emph{direction} (from left to right or from right to
left). The centroid is chosen uniformly, and the length and distance are
chosen from given distributions (see below). Note that some 
centroid, distance, and length combinations are invalid; those
combinations are rejected. Similarly, a \emph{deletion} removes a
portion of the sequence, and can be characterized by a \emph{centroid}
and a \emph{length}. Again, some combinations will be invalid and
are rejected. Each event is a deletion with probability $P_x$, and
a duplication with probability $(1-P_x)$. This process straightforwardly
defines the probability $P(E\,|\,\mathrm{len})$ of any duplication or deletion 
event $E$. Here,
$\mathrm{len}$ is the length of the sequence just before the event
$E$.
The number of events on each branch is governed by a
Poisson process with rate $\lambda$, and thus the probability of
observing $k$ events on a branch of length $\ell$ is $P_n(k,\ell) =
(\lambda\ell)^{k}e^{-\lambda\ell}/k!.$ 

A duplication history $H$ generated in this way implies a set $\sigma(H)$ of
\emph{atomic segments} of several types, as defined in the previous
section. For each type $x$, the history also implies a segment tree
$T_x$. The substitutions in the nucleotide sequences of atom $x$ 
are governed by the HKY substitution model along the corresponding
segment tree $T_x$.

We can compute the joint probability $p(H,X)$ of a given set of extant
sequences $X$ and a history $H$ (up to a normalization constant) as
follows. Let $T$ be a species tree with branches $b_1,b_2,\dots$
Then:
\begin{equation}
p(H,X)\propto \prod_{b_i\in T} P(H,b_i) \times \prod_{x\in\sigma(H)} P(X_x\,|\,T_{x}),
\end{equation}
where $P(H,b_i)$ is the probability of events of history $H$
that occur on branch $b_i$ of the species tree, $X_x$ represents nucleotide
sequences of atoms of type $x$, and
$P(X_x\,|\,T_{x})$ is the probability of these sequences given tree $T_{x_i}$. 
For a sequence of
events $E_1,\dots,E_k$ on branch $b_i$, the probability
$P(H,b_i)$ is simply:
\begin{equation}
P(H,b_i) = P_n(k,\ell)\prod_{j=1}^{k} P(E_j\,|\,\mathrm{len}(j-1))
\end{equation}
where $\mathrm{len}(j-1)$ is the length of the sequence before event $E_j$.

To reduce the number of model parameters, we use
geometric distributions to model lengths and distances of duplication
events. To estimate these distributions, we have used the lengths and
distances estimated by \citet{Zhang2008} from human genome gene
clusters (mean length 14,307, mean distance 306,718). The geometric
distributions seem to approximate the observed length distributions
reasonably well. Similarly, we estimated the probability of inversion
$P_i = 0.39$ from the same data, we set the probability of deletion as
$P_x=0.05$, and the length distribution of deletions matches the
distribution of duplication lengths.

Note that for our application, the normalization constant for $p(H,X)$
does not need to be computed. We assume a uniform prior on length
distribution of ancestral lengths. This has only a small effect for fixed
extant sequences, since the ancestral length is determined mostly by the
length of individual atomic segment types, since the ancestral
sequence should contain one occurrence of each segment type. 
Some combinations of centroids, distances, and lengths will be
rejected, but we assume that in long enough sequences, the effect of this
rejection step will be negligible and we ignore it altogether. 

In the MCMC algorithm below, we compute likelihood 
$P(X_x\,|\,T_x)$ and branch lengths for each 
segment tree separately.
This independence assumption simplifies computation and
allows variation of rates and branch lengths between atoms. This
is desirable, since sequences of different functions may evolve at
different substitution rates, and selection pressures may change the
proportions of individual branch lengths. On the other hand, 
branch lengths tend to be correlated among segment trees
when individual atoms are duplicated together, and this
information is lost by separating the likelihood computations. We are
working on a more systematic solution to the problem of rate and branch
length variation.

\section{Metropolis--Hastings sampling}

We use the Metropolis--Hastings Markov chain Monte Carlo algorithm
\citep{Hastings1970} to sample from the posterior probability 
distribution $p(H\,|\,X)$
defined in the previous section, conditional on the extant sequences $X$
and their atomization. The result of the algorithm is
a series of samples that can be used to estimate expectations of
quantities of interest (e.g., numbers of events on individual
branches, posteriors of individual segment trees, and particular
ancestral sequences), or
to examine high likelihood histories. 

Briefly, the Metropolis--Hastings algorithm defines a Markov chain whose
stationary distribution is the target distribution, but the moves of
the Markov chain are defined through a different \emph{proposal
distribution}. We start by initializing sample history $H_0$.
In each iteration, we use a randomized \emph{proposal algorithm} described
below to propose a candidate history $H'_i$ according to a
distribution conditional on sample $H_{i-1}$. Sample $H'_i$ is either
\emph{accepted} ($H_{i}:=H'_i$) with probability $\alpha(H_{i-1},H'_i)$, 
or \emph{rejected} ($H_{i}:=H_{i-1}$) otherwise. The acceptance probability
$\alpha(H,H')$ is used to ensure that the stationary distribution of
the Markov chain is indeed the target distribution \citep{Hastings1970}:
\begin{equation}
\alpha(H,H') = \min\left(1,\frac{p(H'|X)q(H\,|\,H')}{p(H|X)q(H'\,|\,H)}\right),
\end{equation}
where $q(H'\,|\,H)$ is the probability of
proposing history $H'$ if the previous history was $H$.

In the rest of this section, we describe the proposal algorithm.  
Even though this algorithm employs a number of heuristics to
improve overall performance of the sampler, 
it only affects the mixing rate and convergence properties of the sampling, 
not the asymptotic correctness of the MCMC algorithm.

The algorithm starts by sampling an unrooted \emph{guide tree $T_x$} for every
atom type $x$. The segment trees implied by the proposed history will
be rooted and refined versions of these guide trees.
Each individual guide tree $T_x$ is sampled from the posterior distribution
of the trees 
conditional on a fixed multiple alignment of all instances of atom
type $x$. By sampling the guide trees, instead of fixing a particular
segment trees (as is done by \cite{Ma2008}), 
we account for uncertainty in the tree topologies. Since the
branches with a small number of expected substitutions cannot be usually
estimated reliably, we collapse those with fewer than 5 expected
substitutions over the length of the atom sequence. 
Thus, the guide trees for shorter atoms,
where uncertainty is high, will be close to uninformative star
trees, while the trees for longer atoms will remain more resolved.

The proposal algorithm then samples a history consistent with the given set
of guide trees.  It does so by starting at the leaves of the trees and
progressively sampling groups of atom pairs to merge, 
until the roots of the trees
are reached and only a single copy of each atom remains.
Merging of two groups of atoms corresponds to unwinding one duplication.  
To obtain a valid history consistent with the guide trees,
the two groups must satisfy several conditions. First,
each of the two groups has to be a contiguous subsequence
of the current atomic sequence. Also, the corresponding
atoms of the two groups must be of the same type. Finally,
the corresponding atoms must
be cherries in their guide trees. (The leaves
$x_i$ and $x_j$ are cherries in $T_x$ if they have the same parent.)

For example, if the most recent duplication copied atoms $x_iy_k$ to atoms
$x_jy_\ell$ then $x_i$ and $x_j$ must be cherries in the tree $T_x$,
and $y_k$ and $y_\ell$ must be cherries in the tree $T_y$. 
Unwinding of this duplication will correspond to removal of $x_j$ and
$y_\ell$ from the trees $T_x$ and $T_y$ and from the atomic sequence. 
Now, the same conditions can be
applied to the second latest duplication. In this way, a particular
set of guide trees can significantly restrict the set of possible
histories.  

The sampling
distribution over candidate groups of atoms is determined by a series of
heuristic penalties described below.  
The multiple alignment for each segment type is created by MUSCLE
\citep{Edgar2004}. Even though it is possible to sample multiple
alignments to prevent potential alignment errors from propagating
throughout the whole analysis \citep{Holmes2001}, such sampling is by itself
computationally intensive. Given that in this paper we
consider sequences of greater than 90\% similarity, we do not expect
multiple alignments to be a major source of error in our reconstructions.
Trees, branch lengths, and HKY nucleotide
substitution model parameters are sampled by MrBayes
\citep{Ronquist2003} with uniform prior over tree topologies, and default
priors for the other parameters. For each segment type $x$,
all the tree samples are precomputed in a run of 10,000 iterations with a
burn-in of 2,500 samples, keeping every 10th sampled tree. In every
iteration of the history proposal algorithm, we keep the previous tree with 95\%
probability, otherwise we choose a new tree randomly from the
pre-computed samples.

\smallskip\noindent
{\bf Proposal  distribution.}
As described above, the proposal distribution for histories is defined by a
sequential sampling procedure that selects groups of atom pairs to merge in each
step.  
The goal is to define this
distribution so that the overall proposal distribution is as close as
possible to the actual conditional distribution $p(H_i'\,|\,H_{i-1})$,
making the acceptance probability as close as possible to one.  Directly
characterizing $p(H_i'\,|\,H_{i-1})$ appears to be difficult, so we settle
for a heuristic weighting function in the proposal distribution for merges
that is designed to produce reasonably good proposed histories.  The
Metropolis-Hastings algorithm will ensure that the retained samples will
accurately reflect the posterior distribution, once the Markov chain
reaches stationarity. 

In each step, we consider all possible duplications consistent
with the current set of guide trees, as well as selected deletion and
speciation events. Deletions do not leave
observable sequence traces in extant species, and thus it is not
possible precisely date them; instead, in the proposal algorithm we
associate deletions with the speciation or duplication events that
occurred before them. 
We allow a single deletion following a duplication. We consider only 
deletions completely inside the source or target sequence of 
the duplication. A speciation event is represented as a 
copy of all atomic segments from
one species to a previously empty sequence of another species,
possibly followed by several deletions in both species.
We only allow
speciations in the partial order imposed by the species tree.  
Additionally, we propose 
only speciations that maximize the total sequence length of matched atomic
segments between the two species. As in the case of duplications, only
segments that are currently cherries in the corresponding guide trees
can be matched. For example, if we have sequences in two species
$S_1=a_1 b_1 c_1$ and $S_2=a_2 b_2 c_2$, and $b_1$ and $b_2$ are not
cherries in the segment tree, we have to propose speciation from an
ancestral sequence $ab_1b_2c$ or $ab_2b_1c$, followed by one deletion
in species $S_1$ and one deletion in species $S_2$. Proposals
that obey these constraints
can be easily generated by a simple dynamic programming algorithm, 
and in the case of many possible
speciation proposals, we only keep 20 highest weight candidates.
Note that it is always possible to propose at least one event
until we reach an ancestral sequence of unique atoms.

\def\mypar#1{\par\noindent\emph{#1}}

We characterize each proposed event by a feature vector
$f_1,\dots,f_k$ and the probability of choosing the event will be
proportional to $\exp(\sum_i w_i f_i)$ for some fixed set of weights
$w_i$. In the rest of this section we briefly describe these features 
and their weights.

\mypar{Target length.} The basis of the weight is the length $\ell$
of duplication or speciation, i.e. how much sequence is removed by
unwinding the event. We set $f_1
= \ln(\ell)$ and $w_1 = 1$.

\mypar{Previously seen event.} To keep the newly proposed 
history similar to the previous sample $H_{i-1}$, we add bonus to events seen
in $H_{i-1}$. This is achieved by a binary indicator feature
$f_2$ and weight $w_2=\ln(10)$.
Some events may not be possible in the new
history due to changes in the guide trees.

\mypar{Branch length mean and variance.} For a given duplication
consistent with the guide tree set, we can compute the mean distance $\mu$
of corresponding cherries in the guide tree (weighted by the lengths
of atoms in nucleotides), and also variance on such distance $\sigma$.
The lower $\mu$ indicates likely more recent events, while large variance
$\sigma$ would indicate that we are merging two or more events that happened
at different times. We set $f_3=\mu$, $f_4=\sigma$, $w_3=-10$, $w_4=-1$.

\mypar{Partial duplication penalty.} If the proposed duplication
is a subset of a larger duplication, 
we set indicator $f_5=1$ and use $w_5=-\ln(100)$.

\mypar{Breakpoint reuse penalty.}
Although, we allow breakpoint reuse,
we favor duplications with fewer breakpoint reuses which seems to
be particularly useful for determining correct direction of duplications.
We have implemented the three conditions stipulated by
\citet{Zhang2008} based on collapsibility of atom pairs
on boundaries of the duplicated segments. We set $f_6$ to the number of 
violated conditions and $w_6=-\ln(10)$;

\mypar{Pair reduction bonus.} Consider the number $\pi$ of distinct pairs of
adjacent atom types that occur in the current set of sequences.  For
input with $n$ atom types, $\pi=n-1$ when we reach the
ancestral sequence, and each duplication reduces $\pi$ by at most
2. This gives us a lower bound on the number of events necessary to
reach the ancestral sequence. We set $f_7$ to be the reduction of $\pi$
achieved by the event ($f_7$ can be negative if $\pi$ 
increases) and $w_7=\ln(10)$.

\mypar{Deletion penalties.}  Deletion associated with a duplication is
penalized by setting $f_8=1$ and $w_8=-\ln(10)$.  In
addition, we penalize longer deletions by setting
$f_9=\ln(d/(d+\ell))$ and $w_9=3$ where $\ell$ is the length of the
target sequence in the duplication, and $d$ is the length of the
deletion. Each deletion associated with a speciation is
penalized by setting $f_{10}=1$ and $w_{10}=-\ln(1000)$.

\mypar{Heat constants.}
Finally, in some rounds of the MCMC sampler, we want to explore
radically new histories, while in other rounds we want to concentrate
on smaller local improvements. Thus, we exponentiate the final event 
weights to a heat constant, which changes from round to round. In our
experiments, we have used cyclic sequence of heats $(0.5,0.6,1,1.2)$.

\section{Experiments}
\label{sec:simulations}

We have implemented the MCMC sampler described above
and verified its functionality on simulated data. For the simulations, we
have estimated branch lengths and HKY model 
parameters (equilibrium frequencies and transition/transversion ratio)
from the UCSC syntenic alignments \citep{Karolchik2008} of human, chimp,
and macaque on human chromosome 22. The (geometric) distributions for
source lengths and source/target
distances, and the proportion of duplications
with inversion were estimated from the analysis of human gene clusters
by \citet{Zhang2008}. Finally, we set the deletion rate to 5\%
of the duplication rate, and the length distribution of deletions to match
that of duplications.

\begin{table}
\caption{\label{tab:dataoverview}%
{\bf Overview of simulated and real data sets.}}
\centerline{\begin{tabular}{lccc@{\qquad}ccc@{\qquad}c@{\qquad}c@{\qquad}c}
\hline
                 & \multicolumn{3}{c}{\bf rate 200 (slow)} &      \multicolumn{3}{c}{\bf rate 300 (fast)} & PRAME & AMY & UGT1A\\ 
                 &  min    &    max  &   mean     &  min    &   max   &   mean     \\
\hline
Seq. len (kb)    &   91    &  295    &   172      &  120    &  387    &   219      & 1000, 200 & 221, 170 & 210, 210, 250\\
No. atom types   &      15 &      53 &       36   &      39 &      57 &       48   & 39       &  44    & 55 \\
No. duplications &       5 &      24 &       15   &      18 &      29 &       23   & $34.9\pm 0.8$& $23.4\pm 0.8$ & $22.9\pm 0.8$\\
No. deletions    &       0 &       3 &        0.8 &       0 &       3 &        1.1 & $9.4\pm 1.9$ & $15.2\pm 1.9$ & $20.2\pm 1.3$\\ 
Species          &    \multicolumn{3}{c}{H,C,R}     &    \multicolumn{3}{c}{H,C,R}          &  H,R      & H,R       & H,C,O       \\
\hline
\end{tabular}}
\end{table}

We used the simulation to create 20 simulated gene clusters in each
of the following two categories: slow
evolving and fast evolving (duplication rate at 200 and 300 times
substitutions per site, respectively). We have applied the algorithm to
atomic segments derived from the simulation. 
However, to emulate the increase in
breakpoint reuse due to
imperfect identification of alignment boundaries in real data sets, 
we have removed
short atomic segments ($<500$bp). A summary
of the resulting data sets can be found in Table \ref{tab:dataoverview}.

\begin{figure}
\begin{minipage}[b]{0.5\textwidth}
\includegraphics[width=\textwidth]{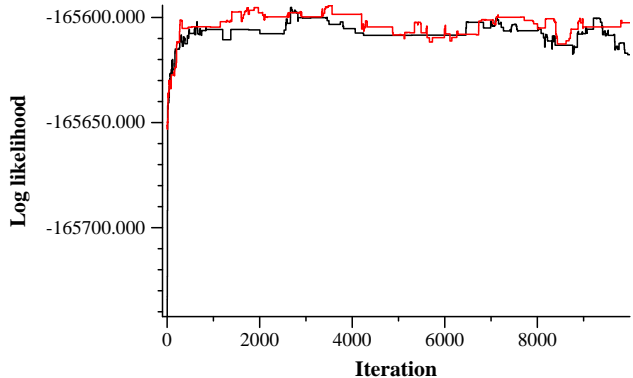}
\end{minipage}
\hfill
\begin{minipage}[b]{0.45\textwidth}
\caption{\label{fig:convergence} {\bf Convergence of the MCMC sampler.}
Log likelihood as a function of iteration number for two independent chains
with random 
starting points on a slowly evolving simulated cluster.}
\end{minipage}
\end{figure}

For each cluster, we ran two chains of the MCMC sampler from 
a random starting points for up to 10,000 iterations each, discarding the
first 2,500 samples as burn-in. The sampler seems to converge reasonably
quickly, as illustrated by Fig.\ref{fig:convergence}.

\begin{table}
\caption{{\bf Performance evaluation along the human lineage.} The table
shows the histogram of differences between the real number of events
and the predicted number of events along the human lineage 
on the 40 simulated data sets
(20 with slow duplication rate, 20 with fast duplication rate).
MCMC: rounded expected number of events from all samples. 
ML: highest likelihood sample. 
Z2008: results by the program of \citet{Zhang2008}.\label{tab:r1}}
\begin{center}
\begin{tabular}{lccccc|ccccc}
\hline
& \multicolumn{5}{c}{\bf rate 200 (slow)} & \multicolumn{5}{c}{\bf rate 300 (fast)}\\
Method & $<0$ & $0$ & $1$ & $2$ & $>2$ & $<0$ & $0$ & $1$ & $2$ & $>2$\\
\hline

MCMC &    1 &  15 &   1 &   3 &   0 &   1 &  16 &   1 &   2 &   0 \\
ML    &   1 &  15 &   1 &   2 &   1 &   1 &  16 &   0 &   3 &   0 \\
Z2008 &   3 &  14 &   2 &   1 &   0 &   1 &   6 &   4 &   3 &   6 \\

\hline
\end{tabular}
\end{center}
\end{table}

A summary of the results on the simulated data is shown in Tables
\ref{tab:r1},\ref{tab:r2}. In the majority of cases we predict the correct
number of events along the human lineage (16 out of 20 for slowly evolving,
16 out of 20 for fast evolving clusters).  
Note that in
some cases the predicted number of events is lower than the actual number
of events: this is likely due to events that become invisible in
present day sequence due to subsequent deletions.
Compared to
\citet{Zhang2008}, the performance has improved, especially in the
case of fast evolving clusters (Table \ref{tab:r1}). However, the 
results of the two programs are not directly comparable, since
our program was run on correct atoms with short atoms filtered out,
and the program by \citep{Zhang2008} used its own atomization procedure
which may make errors.

Table \ref{tab:r2} shows the distribution of predicted events along
the individual branches of the phylogeny. In some cases, events are
predicted to occur on the wrong branch of the phylogeny, but the
differences between the predicted and actual numbers of events are
small. 

\begin{table}
\caption{{\bf Distribution of events along the individual branches
of the phylogeny.} The table shows a histogram of the differences
between the actual and the expected number of events computed from
the MCMC samples.\label{tab:r2}}
\begin{center} 
\begin{tabular}{lccccc|ccccc}
\hline
& \multicolumn{5}{c}{\bf rate 200 (slow)} & \multicolumn{5}{c}{\bf rate 300 (fast)}\\
Branch & $<0$ & $0$ & $1$ & $2$ & $>2$ & $<0$ & $0$ & $1$ & $2$ & $>2$\\
\hline
\multicolumn{11}{l}{\bf Duplications:}\\
   human  &   0 &  20 &   0 &   0 &   0 &   0 &  20 &   0 &   0 &   0 \\
  hominid &   1 &  19 &   0 &   0 &   0 &   5 &  15 &   0 &   0 &   0 \\
  chimp &   0 &  20 &   0 &   0 &   0 &   0 &  20 &   0 &   0 &   0 \\
  macaque &   6 &  13 &   1 &   0 &   0 &   2 &  18 &   0 &   0 &   0 \\
     root &   0 &  15 &   5 &   0 &   0 &   0 &  17 &   2 &   1 &   0 \\
    total &   3 &  16 &   0 &   1 &   0 &   4 &  16 &   0 &   0 &   0 \\
\multicolumn{11}{l}{\bf Deletions:}\\
   human  &   0 &  20 &   0 &   0 &   0 &   0 &  20 &   0 &   0 &   0 \\
  hominid &   1 &  16 &   3 &   0 &   0 &   0 &  15 &   5 &   0 &   0 \\
  chimp   &   0 &  20 &   0 &   0 &   0 &   0 &  20 &   0 &   0 &   0 \\
  macaque &   0 &  18 &   2 &   0 &   0 &   1 &  18 &   0 &   1 &   0 \\
     root &   0 &  19 &   1 &   0 &   0 &   0 &  20 &   0 &   0 &   0 \\
    total &   0 &  17 &   1 &   2 &   0 &   1 &  12 &   6 &   1 &   0 \\
\hline
\end{tabular}
\end{center} 
\end{table}

\begin{figure}
\centerline{\includegraphics[width=0.5\columnwidth]{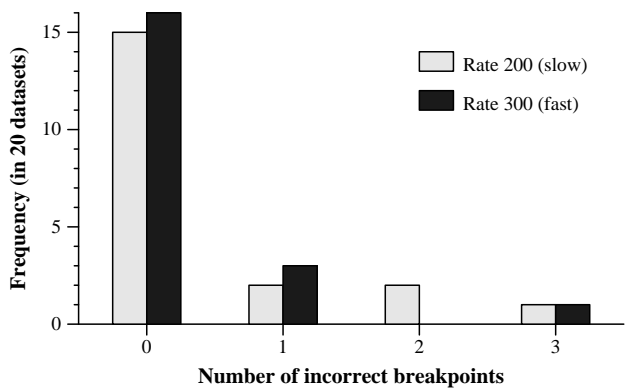}}
\caption{{\bf Histogram of expected number of incorrect breakpoints} 
on the 40 simulated
data sets. The number of incorrect breakpoints is computed for all MCMC
samples and the average is rounded to the closest integer. \label{fig:breakpoints}}
\end{figure}

We have also compared predicted and actual ancestral atomic
sequences (Fig.\ref{fig:breakpoints}). In the vast majority of
cases (31 out of 40), the expected number of incorrect breakpoints
is smaller than $0.5$. 

Beyond the simulated data, we have applied our algorithm to the
following gene cluster sequences: 
PRAME (human-macaque
phylogeny), AMY (human-macaque phylogeny), and UGT1A (human-chimp-orang
phylogeny). The PRAME (preferentially expressed antigen in melanoma)
cluster is one of the most active gene clusters in human genome, and shows 
strong evidence of positive selection \citep{Birtle2005,Gibbs2007}.
The AMY cluster contains five amylase genes that are
responsible for digestion of starch. It appears to have expanded
much faster in humans than in other primates, according to aCGH experiments
\citep{Dumas2007}. The UGT1A cluster consists of multiple isoforms of a
single gene that is instrumental in transforming small molecules into water-soluble and
excretable metabolites. This gene has at least thirteen unique
alternate first exons that result from duplications at various stages
of mammalian evolution.  UGT1A therefore provides an unusual opportunity
for studying promoter evolution. 

\begin{figure}[t]
{\bf UGT1A:}\\[-\baselineskip]
\includegraphics[scale=1]{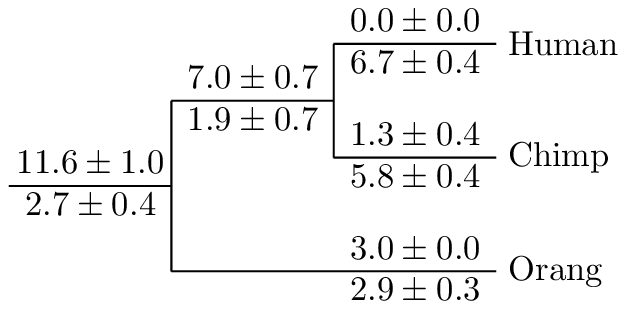}\\[5pt]
{\bf PRAME:}\\
\includegraphics[scale=1]{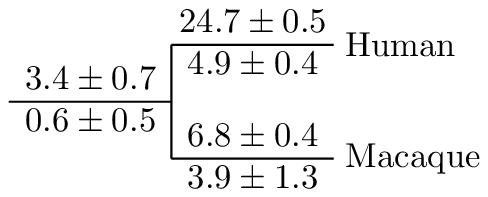}\\[5pt]
{\bf AMY:}\\[-\baselineskip]
\includegraphics[scale=1]{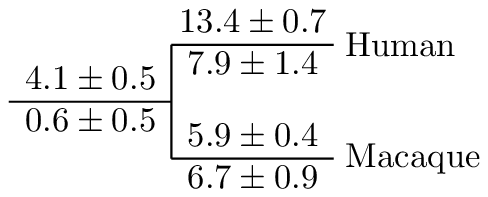}\\[5pt]
\caption{\label{fig:clusters} {\bf Estimated numbers of events.}
For each cluster, we show the posterior mean and standard deviation of 
the number of duplications
(above the branch) and deletions (below the branch) as assessed by MCMC sampling. The root branch
shows events up until 90\% sequence similarity cutoff.}
\end{figure}

Recently duplicated clusters tend to be grossly
missassembled in shotgun-based genomes \citep{Green2001,Zhang2008}, 
so we have to rely on BAC sequencing. We have
screened BACs from chimp, orangutan, and macaque sequenced at
Washington University St. Louis and Baylor Medical College for
similarity with the corresponding human sequences. We
have assembled overlapping BACs into longer contigs and
selected subregions whose ends showed clear homology with upstream and
downstream portions of the human cluster. Then, we have applied a simple
method to 
identify atomic segments.
Briefly, we divide the
sequences into equally sized 500bp windows, and for each window we find 
approximate copies in all available sequences at 90\% 
identity cutoff. The atoms are assigned in a greedy way (starting from
the windows with the largest number of copies), and windows overlapping
already assigned atoms are discarded. Finally, atoms that always occur
in pairs are merged into longer atoms. Table \ref{tab:dataoverview} shows
an overview of the resulting
sequences and atoms.

For each cluster, we ran five chains from
different starting points for 5,000--10,000 samples, discarding
the first 2,500 samples. We have estimated
the number of duplications and deletions overall (Table
\ref{tab:dataoverview}), and on individual branches of the phylogeny
(Fig.\ref{fig:clusters}). The estimated numbers of duplications for
PRAME and AMY are comparable to those of Zhang \etal{}
\citep{Zhang2008}.  With UGT1A, we obtain higher estimates
possibly due to differences in our atomization procedure, and/or
effects of the additional species in the analysis.

\begin{figure}
\centerline{\includegraphics[width=0.4\columnwidth]{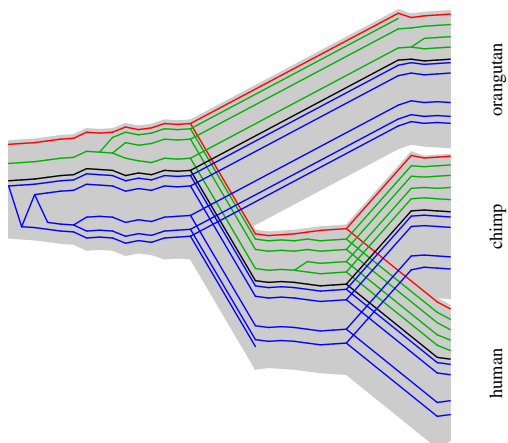}}
\caption{{\bf Highest likelihood reconstruction of UGT1A duplication history.}
The cluster consists of several five exon alternatively spliced isoforms
of UGT1A gene. Exons 2-5 (red) are shared among all the isoforms, while
the first exons (blue, black, green) are alternatively spliced. The
branch lengths in the figure do not correspond to the actual branch lengths.
The atoms are ordered in their order along the genomic sequences (extant and
ancestral).
\label{fig:ugt1a}}
\end{figure}

A tube tree shows the duplication history of several atomic segments
in the context of the species tree, and their locations in 
the extant and ancestral sequences.
Figure \ref{fig:ugt1a} shows the tube tree for the highest likelihood
reconstruction of the history of the UGT1A cluster. 
This cluster
consists of several isoforms of the same five-exon gene. Atoms
corresponding to the protein coding exons are highlighted in color.
Exons 2-5  are shared among all the isoforms, while exon 1 is
alternatively spliced. Our analysis shows clear division of the
first exons into three distinct groups (green, black, and blue) and
their ortholog/paralog relationships in human, chimp, and orangutan.

\begin{figure}
\centerline{\includegraphics[width=0.4\columnwidth]{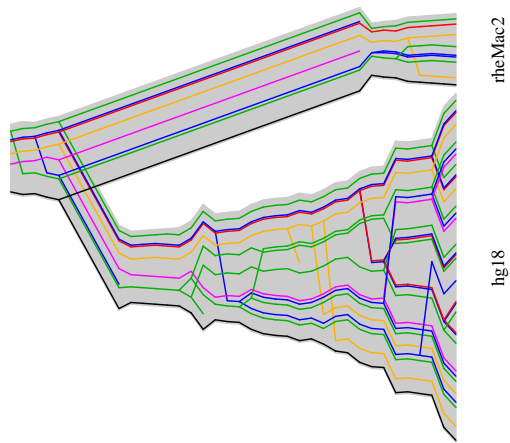}}
\caption{{\bf Highest likelihood reconstruction of PRAME duplication history.}
The cluster consists of multiple copies of the PRAME gene. A typical copy
has three coding exons, the highlighted atoms overlap exon 2. Some
of the genes were pseudogenized. The
branch lengths in the figure do not correspond to the actual branch lengths.
The atoms are ordered in their order along the genomic sequences (extant and
ancestral). \label{fig:prame}}
\end{figure}

While the duplication history of the UGT1A cluster consists of mostly
ancient events, the PRAME cluster (Fig.\ref{fig:prame}) shows very
recent large-scale duplications, especially in the human lineage. In the tube
tree in Fig.\ref{fig:prame}, such events are shown by several
co-linear bifurcations at the same level of the tube tree.
The reconstruction of the evolutionary history of this cluster by traditional
methods (gene tree/species tree reconciliation) is complicated
by the presence of recent duplications (99\% similarity), and presence of the 
chimeric genes \citep{Gibbs2007}. We address these issues
by considering 
multiple guide segment trees for each atom as well as spacial configuration
of atoms in multiple species.
However, this predicted history is by no means perfect.  Rhesus
sequence exhibits a large regions which apparently arose by a single
duplication with reversal; however due to differences in the atomic
sequence this event was split into several shorter duplications.
We expect that improved procedure for segmenting sequence into atoms will
help to address this problem.

\begin{figure*}[b]
\includegraphics[width=0.9\textwidth]{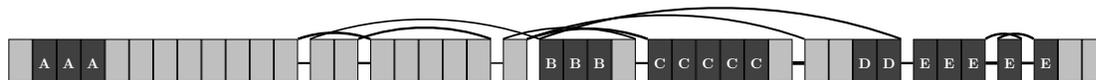}\\
\caption{\label{fig:ancestral} {\bf Ancestral sequence reconstruction for PRAME.}
The cartoon shows large blocks of consecutive atomic segments, with block size
proportional
to the number of atoms per block. The blocks are ordered according
to the highest posterior ordering and the alternative edges show other
possible pairs of adjacent atoms with $>25\%$ posterior probability.
The atoms spanning five ancestral genes at 90\% similarity are marked A-E.
}
\end{figure*}

Fig.\ref{fig:ancestral} shows a
cartoon of the ancestral atomic PRAME sequence of the highest posterior
probability, as well as other pairs of adjacent atoms with 
posterior probability $>25\%$. This reconstruction of
ancestral atomic sequence shows more uncertainty than similar reconstructions
in the simulated data, though there are large blocks resolved uniquely.

\section{Discussion}

In this paper, we have introduced a new model of evolution of
duplicated gene clusters and designed an MCMC-based algorithm that
allows reconstruction of high probability evolutionary histories
of such gene clusters. We have tested our method
on both simulated and real data.

Methods in comparative genomics traditionally concentrate on sequences
where 1:1 orthology can be established. In case of gene clusters, this
is rarely the case, due to their complex evolutionary histories. Our
efforts in reconstruction of gene cluster evolutionary histories will
yield accurate segment trees that will support further development of
comparative genomic tools that can be applied to analyze these complex
regions.

However, gene clusters should not be seen only as a confounding
factor.  The number of orthologous sequences studied, as well as their
divergence and phylogenetic relationships greatly impact the accuracy
of comparative genomic studies. For example, \citet{Kosiol2008} has
shown that the sensitivity of scans for positive selection is greatly
helped by presence of complex phylogeny. While studies based on
orthologous regions between species can provide us with a phylogeny of
up to 10 orthologous copies of a particular mammalian gene at present,
some clusters can provide much more copies with significantly more
complex phylogeny even within a single species (for example, the PRAME
cluster in the human genome contains more than 30 copies). Thus,
the gene clusters provide an opportunity for refined look at
evolution of genes and genomes.
Multiple sources of evidence suggest that many of the
interesting developments in genomes happen within the boundaries of
gene clusters, which further increases our interest in their study.

Multiple efforts are currently under way to recover accurate sequences
of selected gene clusters in multiple species and in multiple populations 
\citep{Zhang2008,Zody2008} by BAC sequencing. Accurate methods
and models for reconstruction of duplication histories of these
clusters are essential to our understanding of evolution, function,
and biomedical implications of these regions.

The general framework of our method allows future developments.  One
limitation of our sampler is its low sample acceptance ratio,
indicating low level of mixing in the Markov chain. We plan to devise
a systematic way for tuning the parameters in the proposal
distribution towards better acceptance ratios.  We also plan to
improve underlying probabilistic model.  Currently the branch lengths
in segment trees are chosen independently of the duplication
history. Instead, we plan to consistently date duplication events on
each branch, and use a scaling parameter for each atom type so that we
can accurately model correlation between branch lengths of individual
atom types and at the same type allow rate variation in different
parts of the sequence. An interesting alternative approach might be to
use combinatorial optimization instead of sampling to find maximum
likelihood history in the above model.

\paragraph{Acknowledgements.} We would like to thank Devin Locke and
LaDeana Hillier at Washington University, St. Louis for providing us
with the BAC sequences from chimp, orangutan, and macaque. 
We would also like to thank Webb Miller and Yu Zhang for helpful
discussions on this problem.

\bibliographystyle{apalike} 
\bibliography{dups}

\end{document}